# CGGAN: A Context Guided Generative Adversarial Network For Single Image Dehazing

Zhaorun Zhou, Zhenghao Shi†, Mingtao Guo, Yaning Feng , Minghua Zhao

School of Computer Science and Engineering, Xi'an University of Technology, No.5 South Jinhua Road, Xi'an 710048, China

**Abstract.** Image haze removal is highly desired for the application of computer vision. This paper proposes a novel Context Guided Generative Adversarial Network (CGGAN) for single image dehazing. Of which, an novel new encoder-decoder is employed as the generator. And it consists of a feature-extraction-net, a context-extractionnet, and a fusion-net in sequence. The feature extraction-net acts as a encoder, and is used for extracting haze features. The context-extraction net is a multi-scale parallel pyramid decoder, and is used for extracting the deep features of the encoder and generating coarse dehazing image. The fusion-net is a decoder, and is used for obtaining the final haze-free image. To obtain more better results, multi-scale information obtained during the decoding process of the context extraction decoder is used for guiding the fusion decoder. By introducing an extra coarse decoder to the original encoder-decoder, the CGGAN can make better use of the deep feature information extracted by the encoder. To ensure our CGGAN work effectively for different haze scenarios, different loss functions are employed for the two decoders. Experiments results show the advantage and the effectiveness of our proposed CGGAN, evidential improvements over existing state-of-the-art methods are obtained.

## 1.Introduction

In haze weather, images captured outdoor usually suffer from serious degradation, such as poor contrast and color distortion [1][2], as illustrated in Figure 1(a). and this will cause great difficulty for further image perception and understanding. Therefore, haze removal is highly desired[3].

To address this problem, a lot of image dehazing methods have been proposed in recent years [4]-[7]. These methods are roughly classified as two categories: image enhancement based methods and image restoration based methods. Image enhancement based methods, such as the histogram equalization[4] and Retinex[5], focus on image contrast enhancement or brightness highlight. Though this kind of methods can improve the visual perception of a degradation image to a certain extent, it fails to address image color distortion and image noise amplification, especially for images with heavily haze, due to do not considering the causes and mechanisms of image degradation.

Different from image enhancement based methods, image restoration based methods such as the Dark channel prior (DCP) dehazing method[6] and the color attenuation prior dehazing method[7], consider the root cause of image degradation, and solve the inverse process of image degradation to obtain a enhanced image. Of which, DCP based method[6], has been given a considerable attention for its effective in most haze scenarios. However, this method has high computational cost due to the use of soft matting for the transmission map refining. To overcome the weakness of He's method, Berman et al. [8] proposed a non-local prior based method. However, this prior can not well describe the haze information in images with complex haze scene or irregular illumination, and lead to inaccurate transmission mapping estimation, thereby fail to restore haze images with sky regions or brightly objects.

✉ Zhenghao Shi
   ylshi@xaut.edu.cn

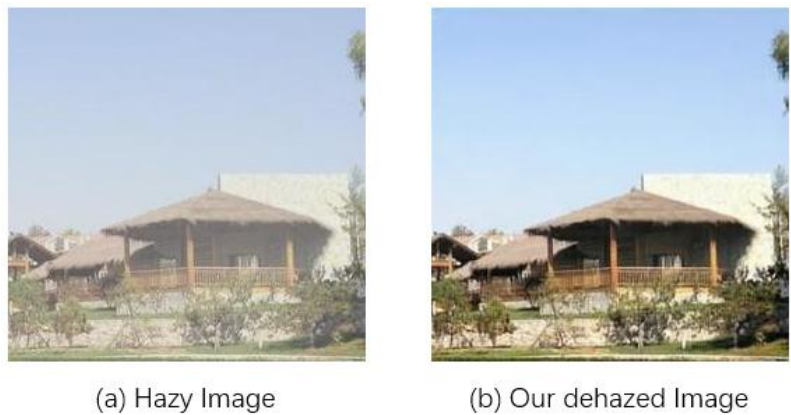

Figure1. An example of CGGAN dehazing.

Deep learning has seen an increasing interesting in image dehazing in recent years[9]-[13]. For example, Cai et al. [9] proposed an end-to-end dehazing network that directly learns and estimates the relationship between haze images and transmission maps. Ren et al. [10] proposed a multi-scale deep neural network to implement image dehazing. Li et al. [11] unify the transmission map and atmospheric light into a formula and directly minimize the pixel domain reconstruction error. Ren et al. [12] proposed a gated fusion network to recover ground true images directly from hazy images. Liu et al. [13] proved that incorporating the atmospheric scattering model into the convolutional neural network would introduce a substantially non-linear component, which would adversely affect the loss surface. Chen et al. [14] used smooth expansion technology and gated sub-network to achieve end-to-end image dehazing.

Though there are varying degrees of success in single image dehazing, almost of all methods mentioned above still suffer from the following limitations:(1) Fail to restore heavily tinted haze images. In heavily tinted haze condition, most lights are scattered and absorbed by atmospheric particles, and causes inaccurate ambient light and transmission estimation. (2) Fail to restore long-shot images naturally, especially for the sky region because of inaccurate estimation of transmission.

To address above mentioned problems, in this paper, we propose a Context Guided GAN (CGGAN) for single image dehazing. In order to make the structure information and content information of the image closer to the real image after image dehazing, we design the generator as a model of one encoder and two decoders. This structure makes better use of the deep feature information extracted by the encoder by adding an extra coarse decoder to the original encoder decoder structure. And multi-scale information obtained during the decoding process of the feature extraction decoder is fused into the fusion decoder, thereby guiding the fine decoder to produce more satisfactory results. We use different loss functions for the two decoders to enable our network to perform well in complex haze scenarios.

Experiments show that the proposed method can obtain good color fidelity, good contrast and proper brightness. Figure 1 (b) illustrates an example of dehazing result with our method.

In summary, our contributions can be summarized in three folds:

1) We propose a new end-to-end CGGAN dehazing network model. This structure makes better use of the deep feature information extracted by the encoder by adding an extra coarse decoder to the original encoder decoder structure.
2) We propose a new encoder-decoder network. By adding an additional decoder, the network guides the decoder to generate higher-quality dehazing results while fully utilizing the encoded features.
3) We use the five loss functions commonly used in current image dehazing to reasonably train CGGAN.

## 2 Related Work

With the success of generating adversarial networks (GANs) and their variants in image processing[15]-[16], a mount of GANs based image dehazing method have been developed these years17]-[21]. For example, Li et.al [17] proposed an end-to-end dehazing network based on CGAN. Inspired by Cycle GAN[16], Deniz Engin [18] proposed a Cycle-Dehaze network by aggregating cyclic consistency and perceptual loss. Zhang at el [19] proposed a novel end-to-end joint optimization dehazing network structure by embedding the atmospheric scattering model into the network. Qu at el [20] turned the image dehazing problem into an image-to-image transform problem. [21]proposes to transform the atmospheric scattering model into a new generated adversarial network. The network can learn the global atmospheric light and the transmission coefficient from data simultaneously and automatically.

For images dehazing, the existing end-to-end dehazing methods mostly use residual blocks or densely connected blocks in the network to obtain clearer dehazing results by increasing the complexity of the network. However, this not only lead a long training time but also cause extremely unstable to optimize. With our further statistic study, we found that many existing deep learning based dehazing methods [10, 11, 18, 19] employ an encoder-decoder as their network architecture. In encoder-decoder, the convolution layer acts as a feature extractor, which captures the abstract content of the image while dehazing. Then use the deconvolution layer to restore the image details. The skip connection connects the convolutional layer and the deconvolutional layer, which can not only make the obtained dehazing result more realistic, but also speed up the convergence speed of training. However, in the case of heavily haze, traditional encoder-decoder cannot well generate real images. The causes of this is that in heavily haze condition, the image details are fuzzed severely, the encoder can not well extract the context information, and thereby the decoder cannot recover the image well due to the lack of information. For what mentioned above, in this work, we introduce a parallel pyramid structure to the traditional encoder-decoder, which not only allows the network to extract more contextual feature information, but also fuses the extracted feature information into the encoder This improvement allows us to get better dehazing result.

## 3 Proposed method

### 3.1 Overview of the Architecture of CGGAN

Figure 2 illustrates the overall framework of the proposed method. It is an end-to-end encoder-decoder generative adversarial network, and consists of a encoder based feature-extraction-net, a parallel pyramid based context-extraction-net, and a decoder based fusion-net in sequence. Of which, the encoder is used for haze feature extraction, the parallel pyramid decoder is used for extracting the context information that is implicit in the deep features. The fusion decoder is used for obtaining a haze-free image by fusing image features obtained by the first two sub-net. We present the details of each sub-net in the following sections.

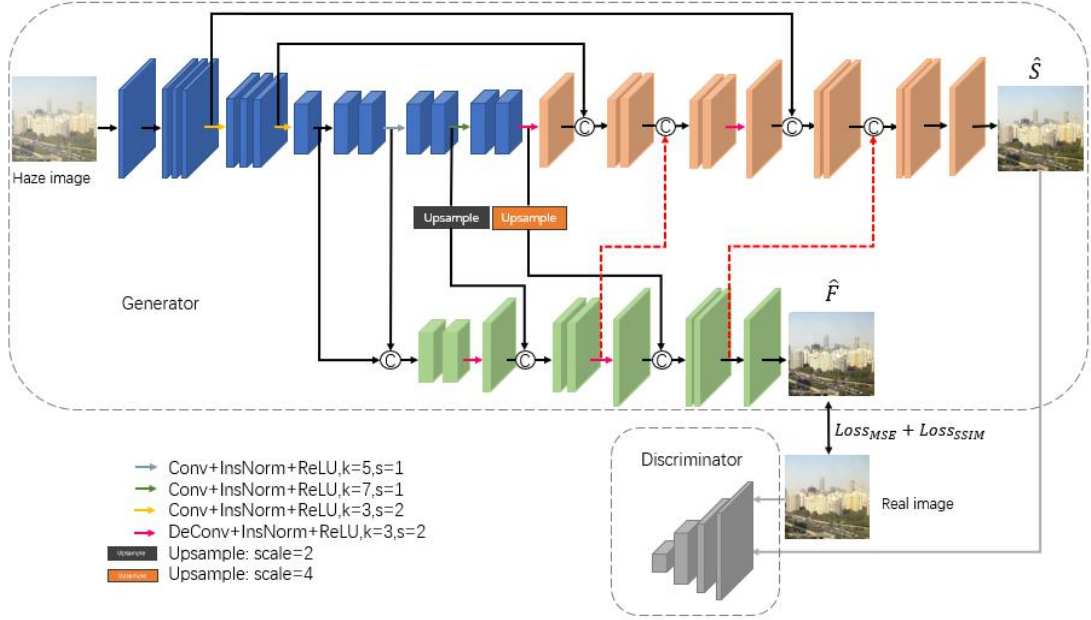

Figure2. The architecture of CGFGAN.CGFGAN includes three parts: the encoder based feature-extraction-net (blue part),parallel pyramid decoder based context-extraction-net(green part) and decoder based fusion-net(orange part).The red dotted line indicates that the skip connection that cuts off the back propagation.

**3.2 Encoder based feature-extraction-net**

In our method, the encoder is divided into two parts, which are used for different purposes. The first part of the encoder takes the haze image as input and extracts the shallow features of the haze image layer by layer, and these features will eventually be combined into the fusion net through skip connections, thereby improving the quality of the dehazed image. The second part of the encoder is a number of cascaded convolutional layers.These convolutional layers, while extracting deep features, serve as the backbone network of the parallel pyramids and provide the necessary feature information for the decoding part of the parallel pyramids.

**3.3 Parallel pyramid decoder based context-extraction-net**

In heavily haze condition, the image details will be fuzzed more severely. Thereby it is very necessary to capture multi-scale context semantic features for scene restoration. Different from existing methods by adding deep feature receptive fields or modifying feature fusion methods to increase context semantic information, we propose a new method to capture the deep features of the encoder (as shown in Figure 3).We use the deep features of the encoder to build a coarse network using parallel pyramids for the backbone network. The main function of the context-extraction-net is to aggregate the context semantic information of the deep features of the encoder and enhance the multi-scale information in a bottom-up manner. The parallel pyramid is divided into two paths starting from the encoded features of the first part. One path is to improve the semantic information by cascading while maintaining the feature size. The second part is a bottom-up feature pyramid model, and the features corresponding to the two paths are fused hierarchically, and a coarse dehaze image with the same size as the input image is obtained. Although this result is not what we ultimately hope to obtain, the multi-scale context information extracted from generating the image will be incorporated into the fine network as a guide feature. While ensuring that the results of the fine network achieve dehazing, the structure and texture of clear images are restored to the maximum.

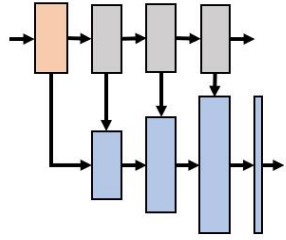

Figure 3. The architecture of Parallel pyramid decoder based context-extraction-net

**3.4 Decoder based fusion-net**

The main role of the decoder is to restore image content details through deconvolution layers. Existing methods only connected the convolutional layer and the deconvolutional layer symmetrically through skip connections. However, as the network becomes deeper, the image details will inevitably be lost. To address this issue, we introduce a parallel pyramid decoder network to make up for the lost semantic information. In the feature fusion process of the encoder, we first use a skip connection to symmetrically connect the shallow features obtained by the convolution layer with the deconvolution layer, and rectify the combined features through the convolution layer. We then fuse the symmetric features in the context-extraction-net into the encoder through skip connections to further compensate the context semantic information lost in image dehazing. On the path that fuses the features of the context-extraction-net to the fusion-net, we set a node to cut back propagation of the two sub-networks. The characteristics of the context-extraction-net only serve as a one-way guide for the refinement of the fusion-net. At the same time, the context-extraction network uses a different loss function than the fine network, which enriches the semantic information of the final dehazing result of the fine network.

**3.5 loss function**

For our proposed CGFGAN, although the context-extraction-net and the fusion-net both decode the features obtained by the same encoder, the focus of the two net is different. the context-extraction-net focuses on the deep features of the encoder, while the fine path not only focuses on the features of the encoder, but also incorporates the features of the context-extraction-net. Therefore, to optimize CGFGAN, different loss functions for the results obtained from two different paths are used. When we use different loss functions for the two paths, the optimization strategy for each path is different, and the decoding characteristics generated during the dehazing process are different. So the feature of the context-extraction-net to the fusion-net can enrich the semantic information and structure information of the fusion-net dehazing result.

**3.5.1 MSE Loss**

The MSE loss is used to optimize the fidelity of the result of the context-extraction-net.The mean square error is the average of the squared sum of the distance context-extraction-net result offset real image. Thus, we optimize the following simple L2 loss function as

$$Loss_{MSE} = \frac{1}{N}(\hat{F} - J)^2 \qquad (1)$$

where $\hat{F}$ and J respectively represent the image dehazing result of the context-extraction-net and the real image.N denotes the total number of pixels.

### 3.2.2 SSIM Loss

Although the dehazing result obtained by the context-extraction-net is not our final dehazing result, the decoding feature map of the context-extraction-net will be transmitted to the fusion-net, so we hope to let the context-extraction-net learn to produce a visually pleasing image. Therefore, we use SSIM loss to improve the image restoration quality of the context-extraction-net .Let x and y are observed and output image respectively.also, G(x) represents output of the proposed generator for the input x.thus, SSIM between G(x) and y is given as follows:

$$\text{SSIM} = [l(G(x),y)]^{\alpha} \bullet [c(G(x),y)]^{\beta} \bullet [s(G(x),y)]^{\gamma} \qquad (2)$$

where, luminance(l), contrast(c), and structural terms(s) are given as,

$$l(G(x),y) = \frac{2\mu_{G(x)}\mu_y + C_1}{\mu_{G(x)}^2 + \mu_y^2 + C_1} \qquad (3)$$

$$c(G(x),y) = \frac{2\sigma_{G(x)}\sigma_y + C_2}{\sigma_{G(x)}^2 + \sigma_y^2 + C_2} \qquad (4)$$

$$s(G(x),y) = \frac{\sigma_{G(x)y} + C_3}{\sigma_{G(x)}\sigma_y + C_3} \qquad (5)$$

if $\alpha = \beta = \gamma$ (the default exponents) and $C_3 = \frac{C_2}{2}$ then Eq.2 reduces to,

$$\text{SSIM} = \frac{(2\mu_{G(x)}\mu_y)(2\sigma_{G(x)y} + C_2)}{(\mu_{G(x)}^2 + \mu_y^2 + C_1)(\sigma_{G(x)}^2 + \sigma_y^2 + C_2)} \qquad (6)$$

where, $\mu_{G(x)}$, $\mu_y$, $\sigma_{G(x)}$, $\sigma_y$ and $\sigma_{G(x)y}$ represent the local means, standard deviations, and cross-covariance for images G(x), y respectively. C1 and C2 the small constants added to avoid the undefined values.

Hence, SSIM loss can be defined as,

$$Loss_{SSIM} = 1 - SSIM(\hat{F}, I) \qquad (7)$$

### 3.2.3 Adversarial Loss

In our fusion-net we adopt the Wasserstein GAN with gradient penalty (WGAN-GP)[27], which instead of clipping network weights like WGAN[28], ensures the Lipschitz constraint by enforcing a soft constraint on the gradient norm of the discriminator's output with respect to its input.so our objective is

$$\text{Loss}(G,D) = \mathbb{E}[D(\hat{S})] - \mathbb{E}[D(J)] + \lambda_{GP}\mathbb{E}_{\hat{x} \sim P_{\hat{x}}}[(\|\nabla_{\hat{x}}D(\hat{x})\|_2 - 1)^2] \qquad (8)$$

where $\mathbb{P}_{\hat{x}}$ is defined as samples along straight lines between pairs of points coming from the ground true data distribution and the generator distribution , and $\lambda_{GP}$ is as weighing factor.

### 3.2.4 MAD Loss

In order to make the image generated by the fusion-net more realistic. We use MAD loss function to capture low level frequencies in the image.The mean absolute deviation loss denotes the $L_1$ loss between fusion-net result and real image . MAD loss is given by

$$Loss_{MAD} = \frac{1}{N}\|\hat{S} - J\|_1 \qquad (9)$$

### 3.2.5 Perceptual Loss

In order to minimize the difference between the perceptual features of the fusion-net result and the real image, we introduce the perceptual loss of the pre-trained VGG-19 network.The formula is as follows:

$$Loss_{VGG} = \frac{1}{C_i W_i H_i} \sum_{c=1}^{C_i} \sum_{x=1}^{W_i} \sum_{y=1}^{H_i} \| \phi(\hat{S}_{c,x,y}) - \phi(J_{c,x,y}) \|_2 \tag{10}$$

where $C_i$, $W_i$, $H_i$ represent the number of channels, width and height of the ith layer feature mapping, respectively. $\phi$ is an operator of nontransformation that extracting a $C_i \times W_i \times H_i$ feature map in VGG-19. (c,x,y) is the pixel position.

### 3.2.6 allover loss function

Therefore, the total loss function of the CGFGAN is:

$$Loss_{CGGAN}^* = \min_G \max_D (Loss(G,D)) + \lambda_1 Loss_{MSE} + \lambda_2 Loss_{SSIM} + \lambda_3 Loss_{MAD} + \lambda_4 Loss_{VGG} \tag{11}$$

## 4 Experiments

In this section, we compare our method with the dehazing results of several state-of-the-art single image dehazing algorithms on indoor and outdoor image datasets, including DCP[6], MSCNN[10], AOD-Net[11], GCA-Net[14], EPDN[20].

### 4.1 Datasets

To verify the effectiveness of our network, our training set consists of indoor and outdoor images. For indoor images, we divided the NYU2 depth dataset image into two parts and randomly cropped it into a size of 256x256x3. we use the method of GFN[12] to synthesize haze images. For training sets, we set the atmospheric light condition A between [0.6-1] and the scattering coefficient β between [0.8-1.6].But for testing sets, we set the scattering coefficient β between [1.0-1.6]. A total of 10000 training datasets and 200 test datasets were generated. For outdoor images, we used the OTS dataset in the RESIDE. We also divided it into two parts and randomly cropped it into a 256x256x3 size to generate 5000 training datasets and 200 test datasets. Thus our training set includes 15000 synthetics image, the test set includes 400 synthetics images, and we guarantee that the training set and the test set image are completely irrelevant.

### 4.2 Training Details

During training, we use Adam optimizer with a batch size of 1, and set a learning rate as 0.0002 ,the exponential decay rates as (β1,β2)=(0.6, 0.999). The learning rate begins to decline linearly when the number of iterations is equal to half of the total number of steps, and decreases to 0 at the end of the network iteration. In order to make the network produce better results, we adjusted the parameters of the loss function. Finally, we found that when $\lambda_1 = 10$, $\lambda_2 = 10$, $\lambda_3 = 100$ and $\lambda_4 = 0.001$, the result is the best. We trained the network for 600000 iterations. We implement our model with the Tensorflow and a RTX 2070 GPU.

### 4.3 Comparisons on Synthetic Test Datasets

Before comparing with the dehazing results of other methods, we first performed a quantitative analysis of the results obtained from context-extraction-net and fusion-net in CGGAN. Figure 4 shows the dehazing results of two sub-networks on indoor haze image and outdoor haze image, respectively. As can be observed that context-extraction-net recovers the structure information in the haze image, but there is still some gap between the color recovery and the haze-free image. However, fusion-net network not only restores the structure information of haze image, but also restores the color information of image.

To quantitative evaluate the performance of our proposed method on the synthetic data, the measures of PSNR (peak signal to noise ratio)[22] and SSIM(Structural Similarity,SSIM) [23] are employed in this experiment. Both of the quantitative measures are calculated using the luminance channel. A higher SSIM indicates greater image structure similarity between a derained image and its ground truth (for the ground truth, the SSIM equals 1.). A high value of PSNR indicates a better dehazing performance.

Table1 shows the PSNR and SSIM values of the two sub network dehazing results. By analyzing the results of the two sub-networks, we verify that the design of passing the decoded features of context-extraction-net into fusion-net can further improve the dehazing ability of the model.

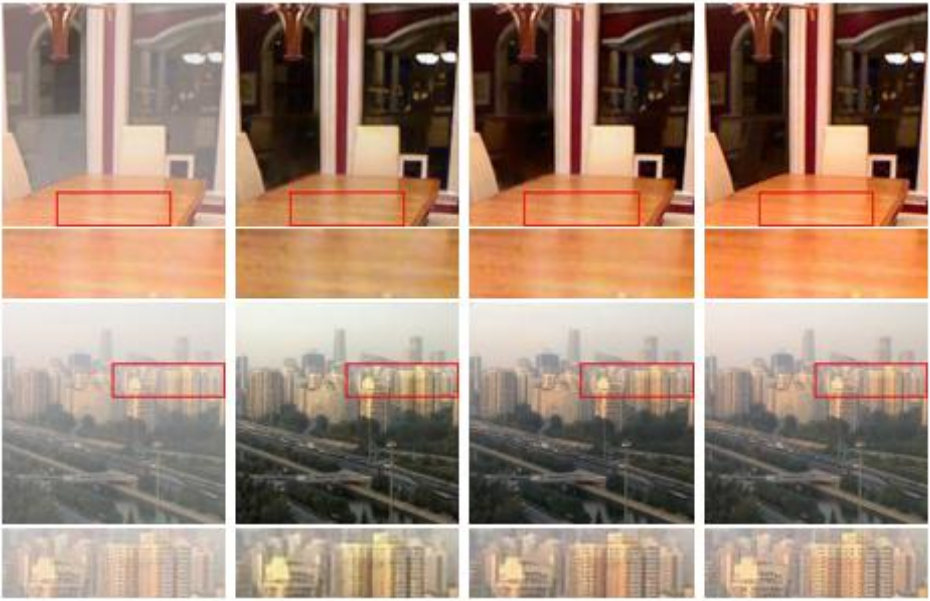

Figure 4. The result of context-extraction-net and fusion-net.

Table 1: Quantitative comparison on different sub-networks.

|  | Metrics | Context-extraction-net | Fusion-net |
|---|---|---|---|
| Indoor | PSNR | 24.13 | **25.17** |
|  | SSIM | 0.8645 | **0.8706** |
| Outdoor | PSNR | 24.99 | **26.42** |
|  | SSIM | 0.8770 | **0.8897** |

To verify the effectiveness of the proposed CGFGAN, we compared it with several state-of-the-art methods on indoor and outdoor test datasets. Figure 5 shows the comparison results of our method and other methods on the indoor image test dataset and outdoor image test dataset. Since we have adopted a haze concentration of the indoor test dataset, we can easily find that our dehazing results have better structure information and semantic information. For the dehazing task of the outdoor test dataset, our model has also achieved very real results. To quantitatively analyze the dehazing results ,Table2 shows the average PSNR and SSIM of dehaze results obtained for each dehaze method on the indoor haze image test dataset and the outdoor haze image test dataset.

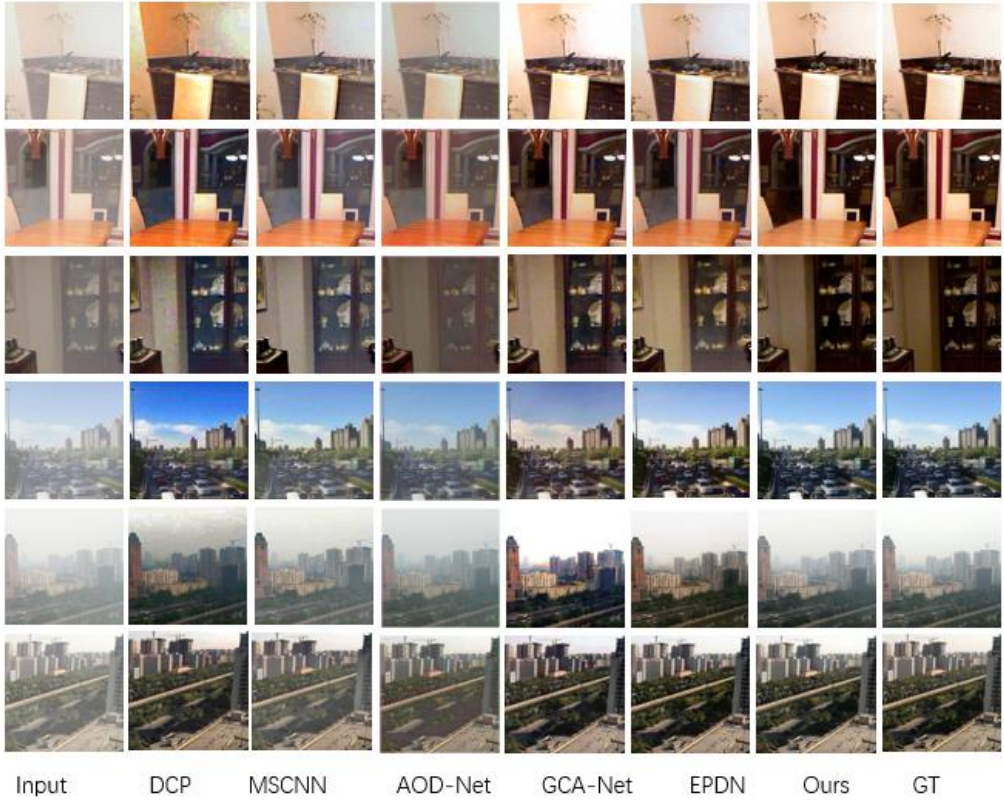

Figure 5: Comparison of the state-of-the-art dehazing methods on the Synthesized Testing Images.the upper three rows indicate the dehazing results of the indoor images and the bottom three rows indicate the dehazing results of the outdoor images.

Table 2: Quantitative comparison on synthetic test datasets

| | Metrics | DCP | MSCNN | AOD-Net | GCA-Net | EPDN | Ours |
|---|---|---|---|---|---|---|---|
| Indoor | PSNR | 19.51 | 19.26 | 17.53 | 22.89 | 21.75 | **25.17** |
| | SSIM | 0.7807 | 0.7692 | 0.7057 | 0.8347 | 0.8307 | **0.8706** |
| Outdoor | PSNR | 19.11 | 19.56 | 19.04 | 20.93 | 21.27 | **26.42** |
| | SSIM | 0.7884 | 0.7993 | 0.7860 | 0.8259 | 0.8237 | **0.8897** |

**4.4 Comparisons on real world images**

To further verify our model's ability to dehaze, we evaluate CGGAN and other dehaze methods on three real-world haze images as illustrated in Figure 6. It can be observed that our CGGAN restores the color and details of haze images well. For the more special sky regions in haze images, CGGAN also completed the greatest degree of recovery.

**4.5 Ablation Study**

In this section, we validate the effectiveness of our proposed objective function, we start with the allocation scheme of the loss function of two paths in the network. We design two schemes: PlanA is to place the adversarial loss, perceived loss, and MAD loss on context-extraction-net, and PlanB is to put the adversarial loss, perceived loss, and MAD loss on fusion-net. From the overall network structure,PlanA actually enhances the dehazing results by adding a sub-network to the GAN, while PlanB adds another sub-network after GAN to optimize the dehazing results.

Figure 7 shows the results of two loss function allocation schemes and Table 3 shows the results of the two loss function allocation schemes quantitatively. We can observe that both methods can obtain very real images, and the quantitative data show that the defogging results are very close.

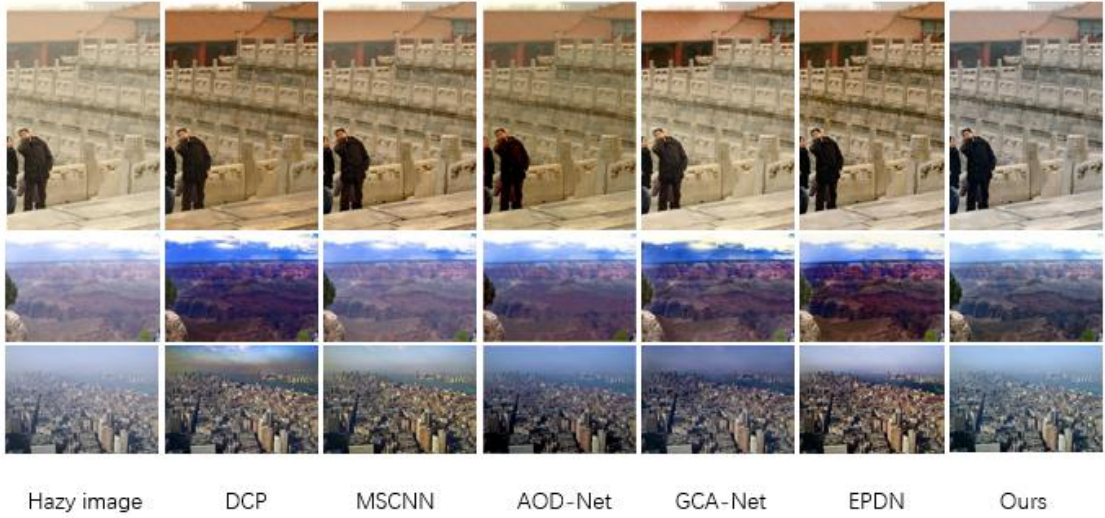

Figure 6: Dehazing results on real-world images

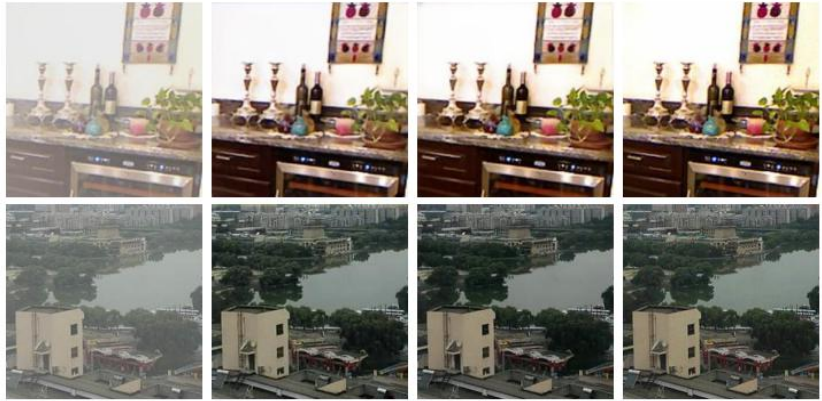

Figure 7 : Dehazing results of different schemes

Table 3: Quantitative comparison on different schemes.

|  | Metrics | PlanA | PlanB |
|---|---|---|---|
| Indoor | PSNR | 25.35 | 25.17 |
|  | SSIM | 0.8865 | 0.8706 |
| Outdoor | PSNR | 25.98 | 26.42 |
|  | SSIM | 0.8932 | 0.8897 |

**5 Conclusion**

In this paper, we propose a new end-to-end single image dehazing network CGAN by improving the traditional encoder-decoder structure. Experiments results show that the new encoder-decoder structure can not only make full use of the features extracted by the encoder, but also perfectly integrate the encoded features into the decoding process, thereby improving the quality of image dehazing. In addition, the CGGAN model also reasonably combines multiple loss functions, making the training of the network more efficient. In the future, our main work is to further improve the network structure, so that our network can deal with more complex haze scenarios.

**Acknowledgment**

The work was jointly supported by the National Natural Science Foundation of China under Grant No. 61872290, and the National Natural Science Foundation of Shaaxi Province under Grant No. 2020JM-463.